# Systematic Evaluation of Long-Context LLMs on Financial Concepts

**Lavanya Gupta[1], Saket Sharma[1], Yiyun Zhao[1]**
[1]Machine Learning Center of Excellence, JPMorgan Chase & Co.,
**Correspondence:** lavanya.gupta@jpmchase.com

## Abstract

Long-context large language models (LC LLMs) promise to increase reliability of LLMs in real-world tasks requiring processing and understanding of long input documents. However, this ability of LC LLMs to reliably utilize their growing context windows remains under investigation. In this work, we evaluate the performance of state-of-the-art GPT-4 suite of LC LLMs in solving a series of progressively challenging tasks, as a function of factors such as context length, task difficulty, and position of key information by creating a real world financial news dataset[1]. Our findings indicate that LC LLMs exhibit brittleness at longer context lengths even for simple tasks, with performance deteriorating sharply as task complexity increases. At longer context lengths, these state-of-the-art models experience catastrophic failures in instruction following resulting in degenerate outputs. Our prompt ablations also reveal unfortunate continued sensitivity to both the placement of the task instruction in the context window as well as minor markdown formatting. Finally, we advocate for more rigorous evaluation of LC LLMs by employing holistic metrics such as F1 (rather than recall) and reporting confidence intervals, thereby ensuring robust and conclusive findings.

## 1 Introduction

Recently, there has been a growing interest in extending the context window sizes of large language models (LLMs) to produce long-context LLMs (LC LLMs) (gpt, 2024; OpenAI, 2024; Gemini Team, 2024). This is especially promising as it allows to extend the "working memory" of LLMs. Real world use cases need LC LLMs to be able to follow increasingly complicated instructions while reasoning over their long context length windows with high degree of reliability.

[1]We do not release the proprietary datasets due to confidentiality concerns.

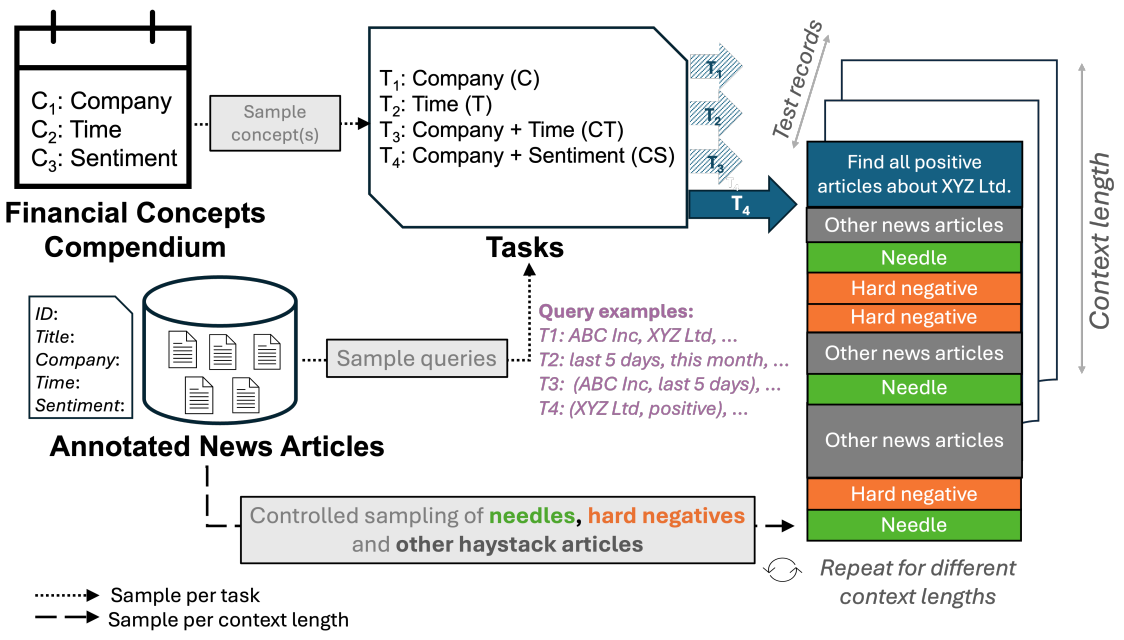

Figure 1: An overview of our framework, made of three financial *concepts* resulting into four real-world practical tasks of varying difficulty levels. This diagram shows an illustration of constructing one test record for task $T_4$: Company+Sentiment (CS). Our framework allows complete control over the sampling and injection of needles, hard negatives and other haystack articles into the context window of an LC LLM.

Current benchmarks test abilities of LC LLMs in several ways. For example, Gemini Team (2024) follow the "Needle-in-a-Haystack" Kamradt (2023); Kamradt et al. (2024) analysis to evaluate Gemini-Pro on retrieval-based long-context synthetic tasks, measuring only surface-level retrieval capabilities. Frequently, there is observed dissimilarity (or heterogeneity) between the needles and haystack in benchmarks like RULER Hsieh et al. (2024), making it artificially easier to retrieve the needle. Other recent work by Wang et al. (2024) includes tasks such as arranging shuffled text segments in the correct order, while FLenQA (Levy et al., 2024) creates balanced True/False datasets inspired by (Weston et al., 2015) to test models' abilities on chaining facts and simple inductions - deviating substantially from real-world usage of LLMs. Other popular long-context benchmarks like L-Eval An et al. (2023), ZeroSCROLLS Shaham et al. (2023), LooGLE Li et al. (2023), LongIns Gavin et al. (2024) and Long-Bench Bai et al. (2023) are inflexible in expanding their text lengths beyond a fixed number of tokens, and hence not applicable to the more recent LC LLMs supporting 128K tokens context window. Finally, metrics like recall remain prevalent Kamradt (2023); Kamradt et al. (2024); Gemini Team (2024); Lee et al. (2024), which however provides only a partial signal on LC LLMs' performance.

[2]The support for tasks T and and CT at 128K context length is slightly different than the other two tasks because tokenization of dates causes the 128K context window to be maxed out, leading to truncation of some articles from the haystack.

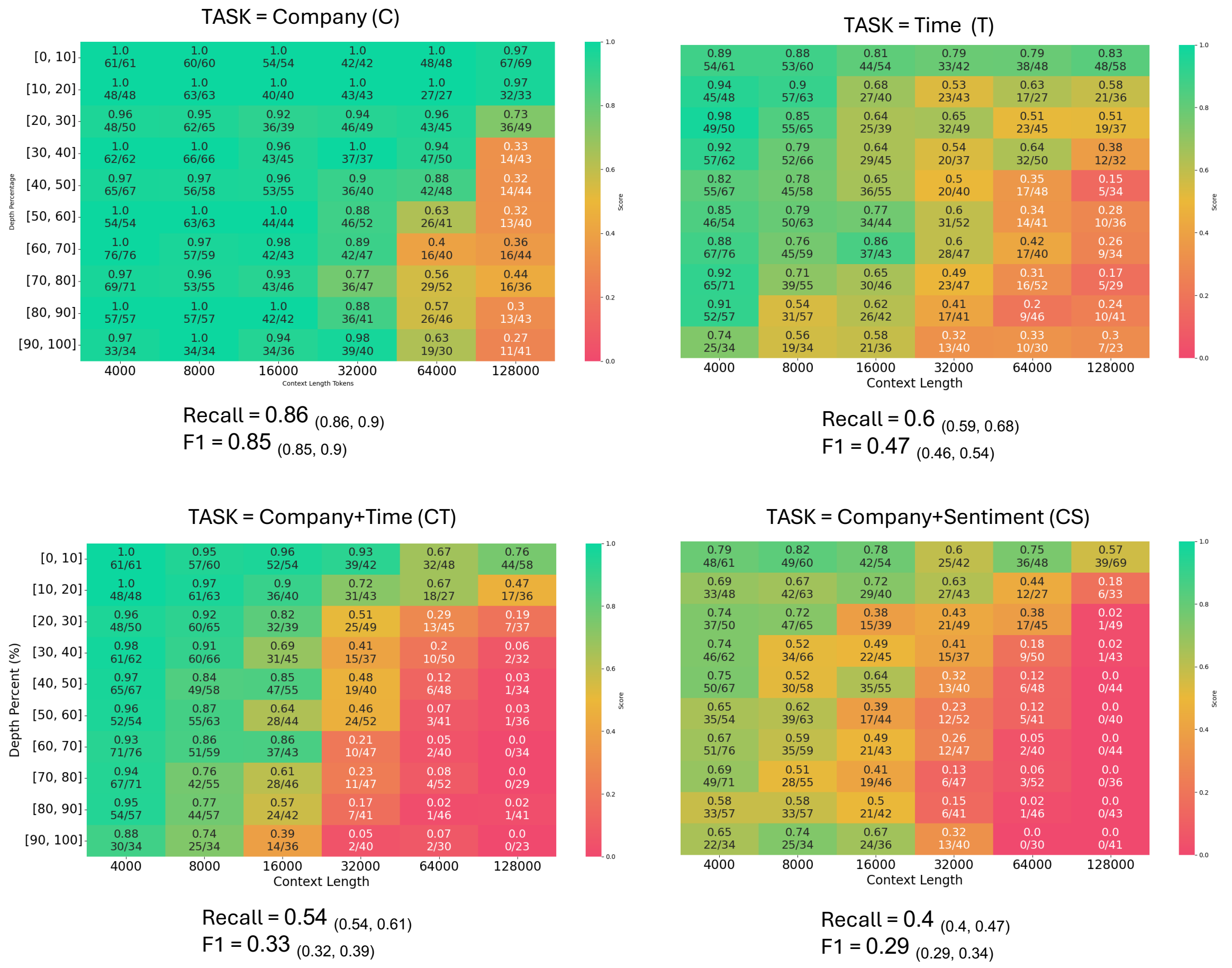

Figure 2: Results of GPT-4o on our benchmark. Each *(x,y)* = (context length, depth) cell is annotated with the computed recall metric. Recall has also been expressed as a raw fraction to clearly indicate the number of correctly retrieved needles (i.e. true positives) vs. the number of gold needles. Aggregated median recall and F1 alongwith 95% CIs across all context lengths are reported below each heatmap. We maintain the same needle support across tasks for fair and consistent comparison[2]. We observe similar trends for GPT-4-Turbo (Appendix A.7).

In this work, we try to address the limitations in existing benchmarks by creating a systematic evaluation framework that is based on real world financial news and tasks of practical interest. Our framework allows the flexibility to manipulate several dimensions crucial for a comprehensive evaluation of long-context LLMs (Table 1). We evaluate two state-of-the-art LC LLMs[3]: GPT-4o (OpenAI, 2024) and GPT-4 Turbo (gpt, 2024), on datasets created using our framework while answering the following research questions:

**RQ1** Does performance depend on the choice of prompting?

**RQ2** Can models reliably use their full context?

**RQ3** Does performance depend on the complexity of the underlying task?

Our findings suggest:

1. Leading LC LLMs are sensitive to both the position of the task instruction as well as minor formatting of the overall prompt. We find that prepending the task instruction to the input context boosts GPT-4 models performance as opposed to appending (Fig. 4).

[3] Refer to Appendix A.6 for exact details on model version and parameters.

2. Models suffer performance degradation at longer contexts on both simple and difficult tasks. For instance, even on the simplest task formulation, the performance declines consistently from F1=0.99 at 4K context length to F1=0.4 at 128K context length (Fig. 2).
3. Model performance deteriorates with increasing task complexity and drops by almost 50 points on simplest vs. most difficult task (Fig 2). Additionally, for difficult tasks, we notice shocking failures in instruction following ability at longer contexts, leading to degenerate model outputs (Fig. 9).

## 2 Methodology

Real world use of LLMs, commonly requires them to locate, reason over and synthesize relevant information across their context window, while accounting for constraints specified in the prompt. Our methodology is inspired to mimic such realistic use-cases of LC LLMs. We now outline the pivotal characteristics of our tasks, experimental setup, prompts and evaluation strategy.

### 2.1 Tasks

We introduce *concepts* that constitute the foundational building blocks of our framework. In this work, we experiment with 3 *concepts*: "companies", "time" and "sentiment" that are relevant in the context of financial news. We combine them as search clauses - single or multiple - to create increasingly difficult long-context retrieval tasks. Unlike other benchmarks that are restricted by the predefined choices of tasks, our framework allows us to flexibly combine concepts to create tasks of varying difficulties, that are a close proxy for realistic tasks.

1. **Company** Company recognition ("Which companies are mentioned in this article?") is a fundamental concept in finance. We covert it into a long-context task that requires LC LLMs to: "Find all articles about <*company_of_interest*> from the articles below."
2. **Time** Temporal queries are frequently encountered in news applications. We create two long-context tasks using this concept: "Find all articles since <*time_range*> from the articles below." and "Find all articles about <*company_of_interest*> since <*time_range*> from the articles below."
3. **Sentiment** Finally, sentiment ("What is the sentiment of <EVENT> in this article?") is another key concept in financial news, that is usually defined in a highly-specialized domain-specific interpretation, inherently making it a more complex concept. Combining with company[6], we convert our 5-class sentiment concept into a long-context task as follows: "Find all <*sentiment_of_interest*> articles about <*company_of_interest*> from the articles below."

In summary, we create combinations of our predetermined financial concepts in curated settings to result into the following 4 functional tasks: Company (C), Time (T), Company+Time (CT), Company+Sentiment (CS).

**Task Difficulty** To provide a background on task difficulty, we share baseline and skyline performance on the underlying concepts (short-context) using zero-shot in-context learning and fine-tuning respectively (Table 2). Firstly, we find that both C and T are relatively simpler concepts for LLMs to understand. We notice that combining concepts results into somewhat harder tasks than its single-concept constituents. Finally, we also note that both C and CS tasks benefit from fine-tuning, showing an improvement of ∼10% and ∼30% over out-of-the-box GPT-4 models respectively. This further suggests that specialized company sentiment classification is a much harder concept for LLMs without fine-tuning. Overall, we observe task difficulty to be a function of the *choice of concept* as well as the *number of concepts* for generalized LC LLMs.

### 2.2 Dataset Creation

We first sample *N=20* unique queries for each long-context task described in Sec. 2.1. Correspondingly, we then perform controlled sampling of news articles from our corpus to create test records of lengths 4K, 8K, 16K, 32K, 64K and 128K tokens respectively. For each of the above context lengths, we feed (116, 233, 462, 950, 1987, 4010) number of articles respectively in the LLM context window. Since every article is annotated with all three concepts, we only sample the haystack context once and re-use it across all tasks enabling us to study model performance as a function of task difficulty independently. In addition, since our entire dataset is sourced from news articles (Fig. 1), we overcome limitations associated with dissimilarity between needles and haystack, making them artificially easy to identify. Refer to Appendix A.1 for more details on dataset properties.

[6] We do not experiment with the "Sentiment" concept alone since entity (company) targeted sentiment is typically more useful than document level sentiment in finance.

| Benchmark | Controlled Context | Task Type | Hard Negatives | Zero-needle |
|---|---|---|---|---|
| NIAH | ✓ | synthetic | ✗ | ✗ |
| RULER | ✓ | synthetic | ✓ | ✗ |
| FLenQA | ✓ | synthetic | ✗ | ✗ |
| BABILong | ✓ | synthetic | ✓ | ✗ |
| MMNeedle | ✓ | synthetic | ✓ | ✓ |
| LongBench | ✗ | hybrid | ✗ | ✗ |
| Ada-LEval | ✓ | hybrid | ✓ | ✗ |
| ZeroSCROLLS | ✗ | realistic | ✗ | ✗ |
| L-Eval | ✗ | realistic | ✗ | ✗ |
| BAMBOO | ✗ | realistic | ✗ | ✗ |
| LOFT | ✓ | realistic | ✗ | ✗ |
| **Ours** | ✓ | realistic | ✓ | ✓ |

Table 1: Comparison between existing long-context benchmarks and ours. A controlled context means the flexibility to completely control the *choice*, *count* and *position* of the key information (or evidence) as well as the content of the context window (haystack) of the LC LLM. A "realistic" task is a meaningful task of practical relevance. "Natural" hard negatives[4] is the reliance on the nature of dataset to contain instances of hard negatives; "Induced" means explicit creation of hard negative samples in the benchmark. Zero-needle is a special edge-case where the needle is absent from the haystack[5].

| **Concept(s)** | **Baseline** | | **Specialized Model** |
|---|---|---|---|
| | GPT-4o | GPT-4-Turbo | |
| Company | 0.79 | 0.79 | <u>0.87</u> |
| Time | 0.97 | 0.89 | - |
| Company + Time | 0.85 | 0.79 | - |
| Company + Sentiment | 0.56 | 0.59 | <u>0.89</u> |

Table 2: We report the F1-scores of out-of-the-box GPT-4 LC LLMs (baseline) against a LoRA fine-tuned Llama-2-7B (skyline). We do not fine-tune for the tasks containing "time" concept due to lack of annotated training data.

Following the "Needle-in-a-Haystack" (Kamradt, 2023) paradigm, for each test record, we randomly sample the number of needles $k \in [1,10]$ to be injected, followed by their randomly sampled positions (or depths) relative to the context length. Overall, each task is evaluated on (110, 110, 85, 85, 85, 85) test records at (4K, 8K, 16K, 32K, 64K, 128K) context lengths respectively, thereby mitigating concerns related to low sample strength.

**Zero needle** In real-world tasks, there is no guarantee that the gold article (needle) will exist in the context. A comprehensive and practical evaluation of LC LLMs therefore, must test the ability of models to handle this scenario appropriately. Therefore, we augment our experiment setup with an equal number of zero-needle test records.

**Hard negatives** Occurrence of hard negatives is a typical natural phenomenon in real-world retrieval tasks. Despite this, the inclusion of hard negatives (or hard distractors) in the evaluation of long-context LLMs has been scarce (Table 1). We address this issue by advisably controlling the hard negatives population to comprise between 10%-20% of the total context length per test record.

In summary, we systematically create controlled tests with varying context lengths, count and placement of needles, percentage of hard negatives, as well as task complexity to test the reasoning and retrieval capabilities of long-context LLMs.

### 2.3 Prompts

We manually craft a prompt for each task, following a generic schema as show in Fig. 3. Prompts are run in zero-shot setting to replicate real-world use of LC LLMs. Outputs are requested as JSONs with an inline example of the expected output structure (Appendix A.4.1).

**Ablation on prompt placement** Multiple works have shown that prompting strategies significantly influence performance. Lee et al. (2024); Li et al. (2024) highlight models' sensitivity to the position of few-shot instances in the prompt, while Levy et al. (2024) study the effectiveness of techniques like Chain-of-Thought (CoT) at longer contexts. In our work, we investigate the impact of

###TASK: From the news articles below, there are multiple articles that have a <sentiment_of_interest> sentiment about <company_of_interest>.

<Definition of underlying concept(s)>

<Output format instruction>

""""""

Article ID: 0

<Title>

Article ID: 1

<Title>

.

.

.

Article ID: n

<Title>

""""""

OUTPUT:

Figure 3: An example prompt from "OpenAI Best practices" configuration for "Company+Sentiment (CS)" task. Each prompt is composed of the task instruction and the input context. We show here the task template, query, concept definition, output format instruction, context and output response marker. All other task instructions can be found in Appendix A.4

varying prompt positions in the context window of a long-context LLM. Specifically, we ablate on the relative position of the instructions with respect to the haystack leading to the following configurations: (1) Prepend: Instructions precede haystack (2) Append: Instructions follow haystack (3) Prepend+Append: Instructions are repeated at both ends of the haystack, and (4) OpenAI Best Practices: Instructions precede haystack alongwith following the specific markdown structure recommended in OpenAI best practices[7].

## 2.4 Evaluation Strategy

In this work, we evaluate two OpenAI state-of-the-art LC LLMs: GPT-4o and GPT-4-Turbo on the four long-context tasks described in Sec. 2.1. We credit scores for both loadable and unloadable JSONs[8].

In our scoring strategy, we measure true positives as the number of ground-truth needles (article IDs) that the model correctly predicted; and false positives as the number of predicted article IDs that were not actually ground-truth needles. We report F1-score on all our tasks in order to provide a complete picture of the model performance. To further ensure robustness, we additionally report bootstrap 95% confidence intervals (DiCiccio and Efron, 1996) that help alleviate concerns with small support prevalent in existing LC LLM evaluations (Kamradt, 2023; Kamradt et al., 2024; Gemini Team, 2024; Aparna Dhinakaran, 2024).

[7]https://help.openai.com/en/articles/6654000-best-practices-for-prompt-engineering-with-the-openai-api

[8]We use the Python utility method *json.loads()* to test loadability. Details on how we handle unloadable JSON outputs is covered in Appendix A.3

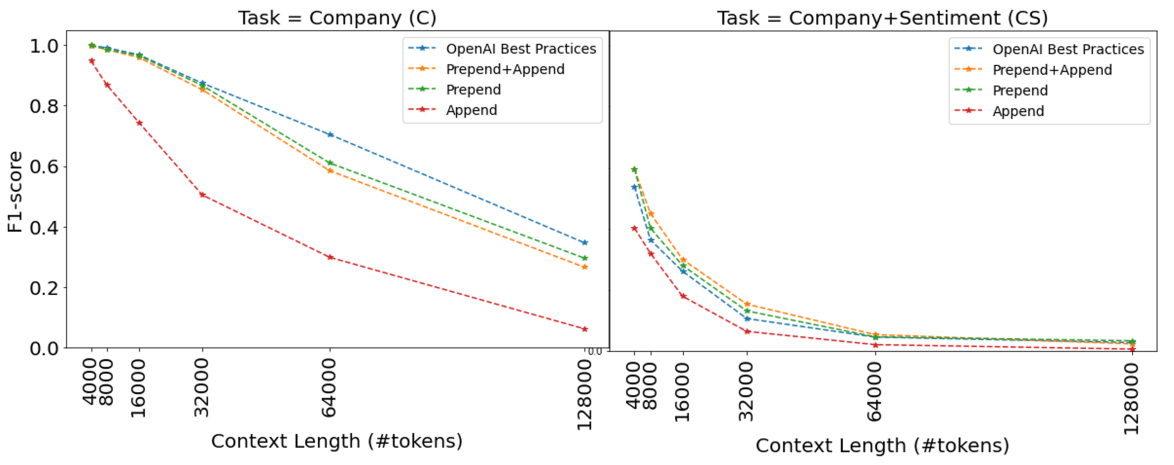

Figure 4: Prompt Sensitivity: GPT-4-Turbo is vulnerable to the placement of the task instruction in the LLM context window.

# 3 Results

To delve into the factors affecting the performance of LC LLMs, our results center on three key research questions presented in Sec. 1.

## 3.1 RQ1: LC LLMs are sensitive to position of the task instruction as well as minor prompt formatting

As discussed in Sec. 2.3, we experiment with four configurations of prompt placement. Our experiments show that while all the three prepend configurations (i.e. "Prepend", "Prepend+Append", "OpenAI Best Practices") are closely comparable, the "Append" configuration is considerably worse (Fig. 4). We also record an overall improvement in performance using "OpenAI Best Practices" configuration over its vanilla "Prepend" counterpart. This exposes the unfortunate sensitivity of state-of-the-art LC LLMs to minor formatting. As a result, all the subsequent results in our work are reported on the "OpenAI Best Practices" configuration.

## 3.2 RQ 2: LC LLMs do not treat all context lengths equally

We observe that LC LLMs do not perform equally reliably at short vs. long context lengths on any given task (Fig. 2). Even on simpler tasks like C, LC LLMs achieve almost a perfect F1-score at smaller context lengths (<= 32K tokens), but start breaking down at longer contexts. This breakdown is observed across all tasks where model performance declines consistently with increasing context length, thereby performing poorly on almost 75% of their claimed context window.

LLMs have been reported to demonstrate a positional bias by utilizing information located at the beginning or end of the input more effectively, commonly termed as the "lost-in-the-middle" (Liu et al., 2024) phenomenon. However, in our experiments, GPT-4o and GPT-4-Turbo do not universally conform to this prevalent assumption across its long context window (Fig. 2). Infact, they exhibit a varying preference towards position of key information at different context lengths and different prompt configurations (Appendix A.8).

### 3.3 RQ3: LC LLMs perform poorly on difficult nuanced tasks

As discussed in Sec. 2.2, our design of (re-)using the same haystack context for all tasks allows us to disentagle and study the model's task ability in isolation to other factors. We observe that for more difficult multi-concept tasks (such as CT and CS), model performance collapses (almost) to 0 at context lengths greater than 32K (Fig. 2), rendering models completely unusable at longer contexts, as also reported by (Wang et al., 2024). However, such stark drops in performance are not observed for relatively simpler single-concept tasks (such as C and T).

**Degenerate outputs** Surprisingly, we notice that LC LLMs start generating degenerate outputs such as repeating themselves or simply counting article IDs in sequence with increasing task difficulty and at longer context lengths (Fig. 5, 6). For instance, for GPT-4-Turbo at 128K context length, we record 4% degenerate responses on task C, and a shocking 42% on task CS. Overall, we observe that GPT-4-Turbo is more vulnerable to such breakdown as compared to GPT-4o. Such failures have also been reported by (Levy et al., 2024). We leave further investigation of this phenomenon to future work.

### 3.4 Zero-needle

Previous works have shown that even powerful LC LLMs are imperfect at rejecting to answer (Zhao et al., 2024). We witness similar behavior in our experiments wherein GPT-4-Turbo successfully returns empty JSONs for easy tasks such as C, but struggle on difficult tasks such as CS (Appendix A.9). We emphasize such tests are crucial to ensure

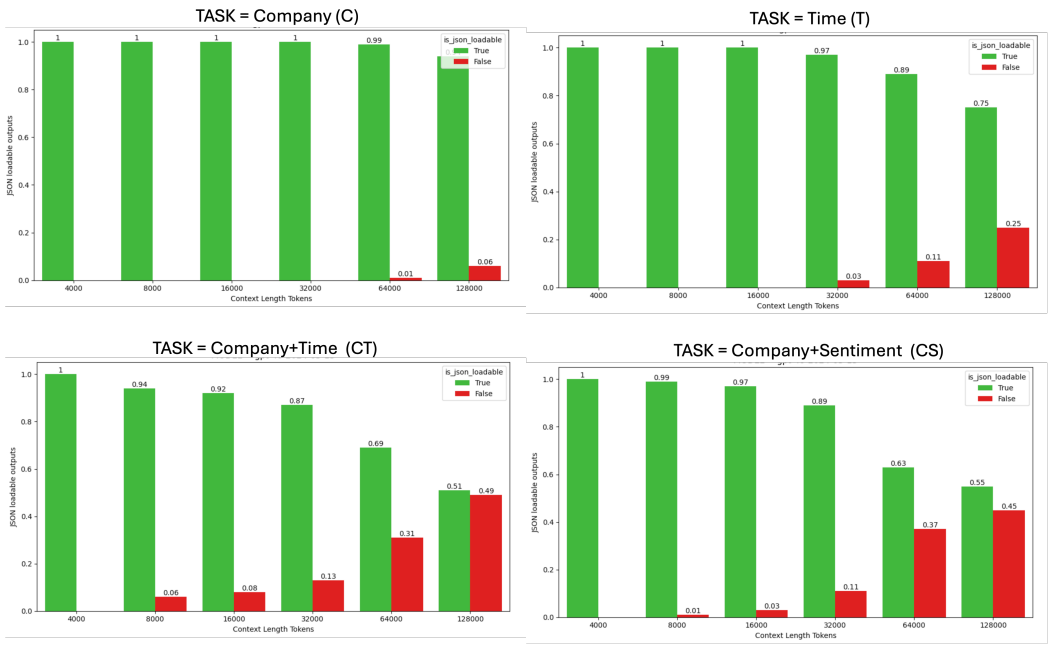

Figure 5: Percentage of invalid JSON model outputs returned by GPT-4o.

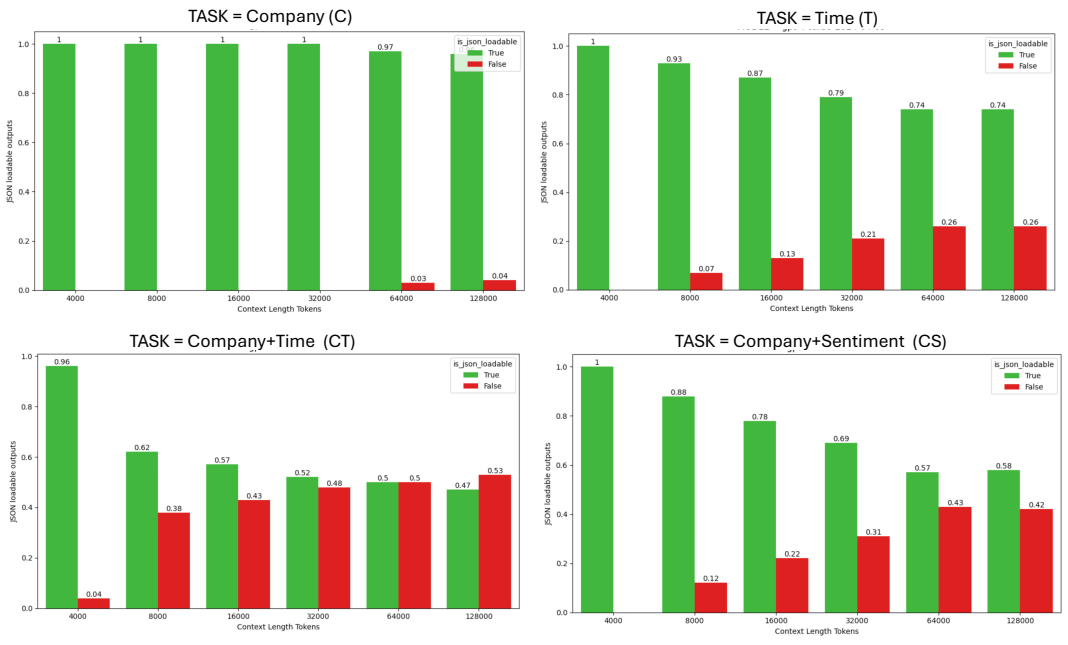

Figure 6: Percentage of invalid JSON model outputs returned by GPT-4-Turbo.

robustness to distracting text in real-world industrial applications.

## 4 Towards Better Evaluation

**Recall is not a reliable metric for difficult retrieval tasks at longer contexts.** Most research in long-context retrieval primarily report their results using recall as the evaluation metric (Kamradt, 2023; Kamradt et al., 2024; Gemini Team, 2024; Aparna Dhinakaran, 2024), since the basic goal is to assess if LC LLMs can "remember" key information in long contexts. We argue that a simple **recall metric is often artificially inflated** (Fig. 7) and hence of limited pragmatic value in real-world systems. For this reason, we also report bootstrap median F1-score throughout this work.

## 5 Conclusion

Our methodical framework characterized by real-world financial news *concepts* allows for flexible configurations to setup a range of different complexity tasks. Our study reveals that long context retrieval and reasoning is still a challenging task for long-context LLMs. Our dataset requires 32K tokens to challenge state-of-the-art GPT-4 models on easier tasks, and only 16K tokens on difficult tasks.

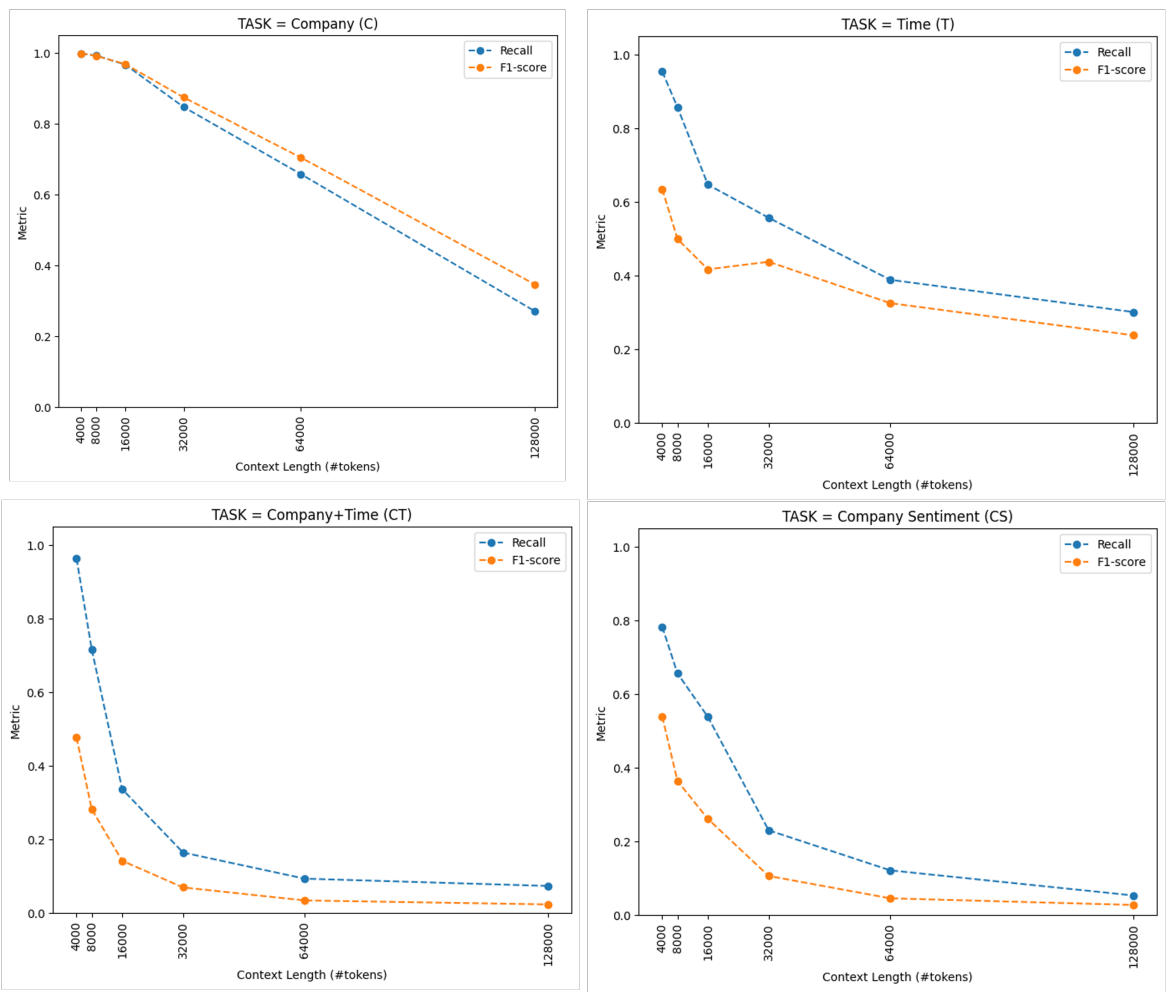

Figure 7: Discrepancy between Recall and F1-score reported on all tasks for GPT-4-Turbo. Each point reflects the average metric of 100 test records.

Models perform poorly on 75% of their context window, with performance declining sharply as input length and task complexity increases. This limitation underscores the need to *reliably* test the effectiveness of long context windows against model provider claims. Moreover, LC LLMs also succumb to minor prompting variations, underpinning the importance of using robust metrics.

## 6 Limitations

In our experiments, we focused on GPT-class models due to organizational constraints, with plans to evaluate other model families in the future. While our tasks are based on real-world scenarios, they do not fully assess the long-form generation capabilities of LC LLMs, which are difficult to evaluate precisely. Additionally, we only begin to explore the complexity of real-world instructions where various constraints are applied. Nonetheless, we hope our evaluations offer valuable insights to guide future research.

## 7 Disclaimer

This paper was prepared for informational purposes by the Machine Learning Center of Excellence (MLCOE) group of JPMorgan Chase & Co. and its affiliates ("JP Morgan”) and is not a product of the Research Department of JP Morgan. JP Morgan makes no representation and warranty whatsoever and disclaims all liability, for the completeness, accuracy or reliability of the information contained herein. This document is not intended as investment research or investment advice, or a recommendation, offer or solicitation for the purchase or sale of any security, financial instrument, financial product or service, or to be used in any way for evaluating the merits of participating in any transaction, and shall not constitute a solicitation under any jurisdiction or to any person, if such solicitation under such jurisdiction or to such person would be unlawful.

# A Appendix

## A.1 Dataset Details

In our dataset, each news article title is about 21 tokens on average, with shortest title having 14 tokens and longest title having 90 tokens. Below are a few (anonymized) example news articles used in this work:

<MASK> receives shares of <MASK> to previously announces spinout

<MASK> in top 2% of price performers of NASDAQ stocks

<MASK> completes acquisition of <MASK>

Can <MASK> continue its gains?

<MASK> bosses buy in after disastrous second half

Investors look to see if <MASK> recent rally will continue

<MASK> stock price passes below 50 day moving average of $0.19

## A.2 Model Failures

A deep dive into model failures revealed two patterns of unloadable JSONs errors: (1) Continuous sequence of generations (2) Repeating their generations indefinitely. An example is shown in Fig. 9.

## A.3 Assigning credits to invalid JSONs

Fig. 8 shows the pseudo-code that we use to handle and parse invalid JSON outputs returned by the models.

```
import re
regex = r'[^0-9,]'
model_response = re.sub(regex, '', model_response)
pred_needle_ids = model_response.split(",")
pred_needle_ids = list(set([int(x) for x in pred_needle_ids if x != '']))
```

Figure 8: Python pseudo-code to parse invalid JSONs outputs

{"article_ids": ["12", "70", "99", "127", "164", "225", "241", "297", "329", "400", "449", "500", "569", "593", "612", "618", "625", "629", "634", "648", "657", "658", "664", "670", "675", "679", "685", "686","692", "693", "694", "695", "696", "697", "698", "699", "700", "701", "702", "703", "704", "705", "706", "707", "708", "709", "710", "711", "712", "713", "714", "715", "716", "717", "718", "719", "720", "721", "722", "723", "724", "725", "726", "727", "728", "",

{ "article_ids": [ "27", "60", "97", "260", "702", "715", "715", "715", "715", "715", "715", "715", "715", "715", "715", "715", "715", "715", "715", "715", "715", "715", "715", "715", "715", "715", "715", "715", "715", "715", "715", "715", "715", "715", "715", "715", "715", "715", "715", "715", "715", "715", "715", "715", "715", "715", "715", "715", ',

Figure 9: Two types of degenerate model outputs

## A.4 Prompts

### A.4.1 Output format instruction

We show below the common output format instruction that we use for all four tasks: C, T, CT and CS.

### A.4.2 Long Context

We show below in Table 3 the four different long-context task templates. The other parts that constitute the full prompt are common to all tasks as shown in Fig. 3 and Fig. 10.

Identify all such articles and return the Article IDs only in a comma separated list in the JSON structure after the OUTPUT marker as follows:
OUTPUT: {"article_ids": <insert list of found Article IDs here>}
For example, if you identify that Article IDs "x", "y", and "z" contain company_name then the output should look like:
OUTPUT: "article_ids": ["x", "y", "z"]
If no Article IDs are found, return the following JSON.
OUTPUT: "article_ids": "[]"
Remember to return all found Article IDs.
Do not give information outside the document or repeat your findings.'

Figure 10: Output format instruction with an inline example of the expected output structure

| Task | Task Template |
|---|---|
| Company (C) | ###TASK: From the news articles below, there are multiple articles which focus on <company_of_interest>. |
| Time (T) | ###TASK: Today is <random_date> specified in YYYY-MM-DD format. From the news articles below dated in YYYY-MM-DD format, find articles published since <time_range>. |
| Company+Time (CT) | ###TASK: Today is <random_date> specified in YYYY-MM-DD format. From the news articles below dated in YYYY-MM-DD format, find articles published since <time_range> that focus on <company_of_interest> |
| Company+Sentiment (CS) | ###TASK: From the news articles below, there are multiple articles that have a <sentiment> sentiment about <company_of_interest> |

Table 3: Task templates for long-context tasks

## A.5 Hard Negatives Examples

Our experiments rely on two types of hard negatives: Natural and Induced. Table 4 explains their definitions with a few examples.

| Concept(s) | Occurrence | Example |
|---|---|---|
| Company | Natural | Similarly named but different companies (eg. ABC, Inc. and ABC, LLC) |
| Time | Natural | (1) Dates that satisfy the query but in a different format (ie. YYYY-DD-MM instead of YYYY-MM-DD) (2) Border-line dates lying just outside the time range query |
| Company + Time | Induced | Same company but different time range |
| Company + Sentiment | Induced | Same company but different sentiment |

Table 4: Hard negatives

## A.6 Model parameters

Model versions and decoding strategy is shared in Table 5. For all our experiments, we set the maximum output generation token length to 100 tokens.

## A.7 GPT-4-Turbo Results

We report results of GPT-4-Turbo on our benchmark in Fig. 11.

## A.8 Prompt Ablations

Fig. 12 shows our prompt ablation results on GPT-4-Turbo. We note that there are no clear trends of "lost-in-the-middle" phenomenon.

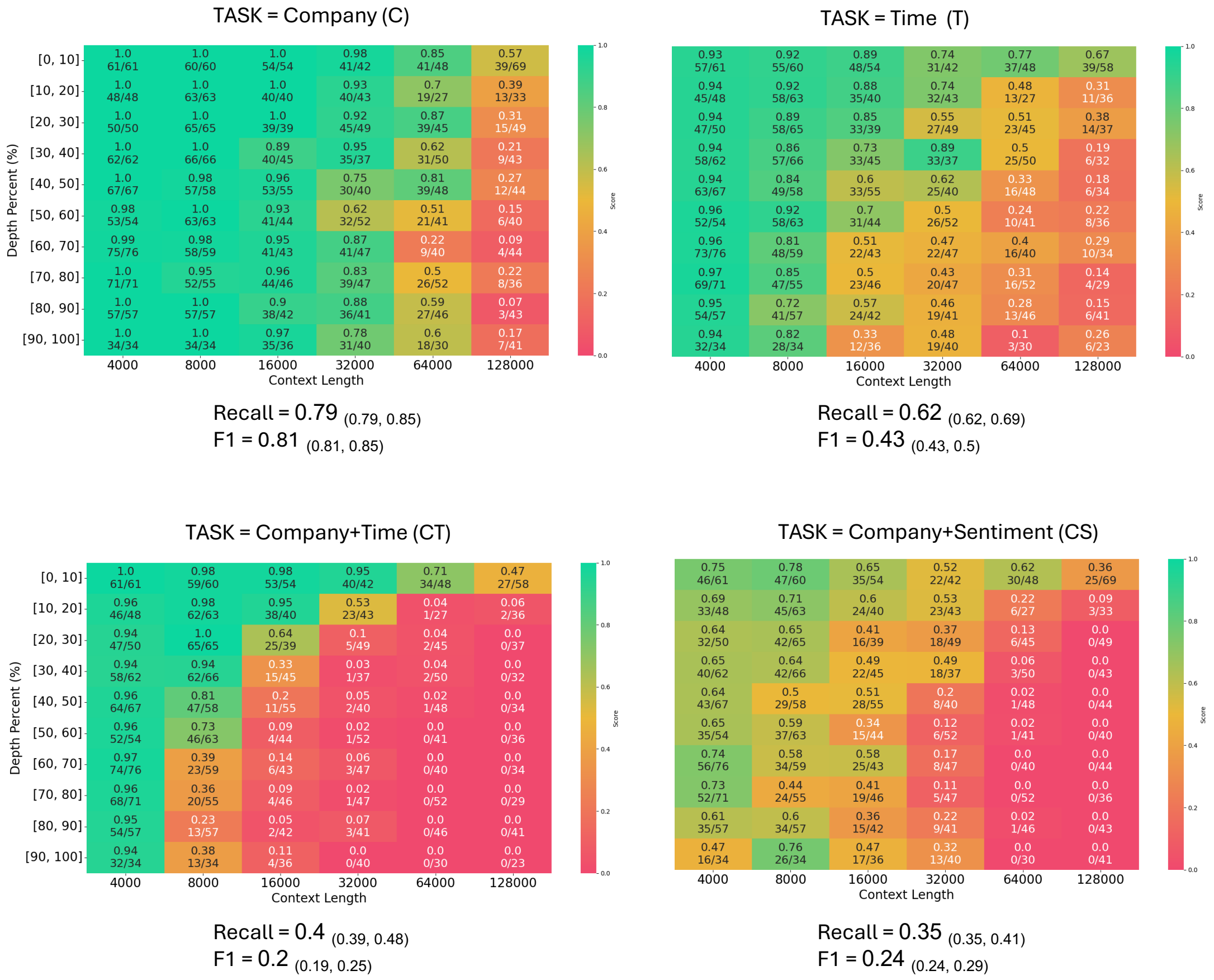

Figure 11: Results of GPT-4-Turbo on our benchmark.

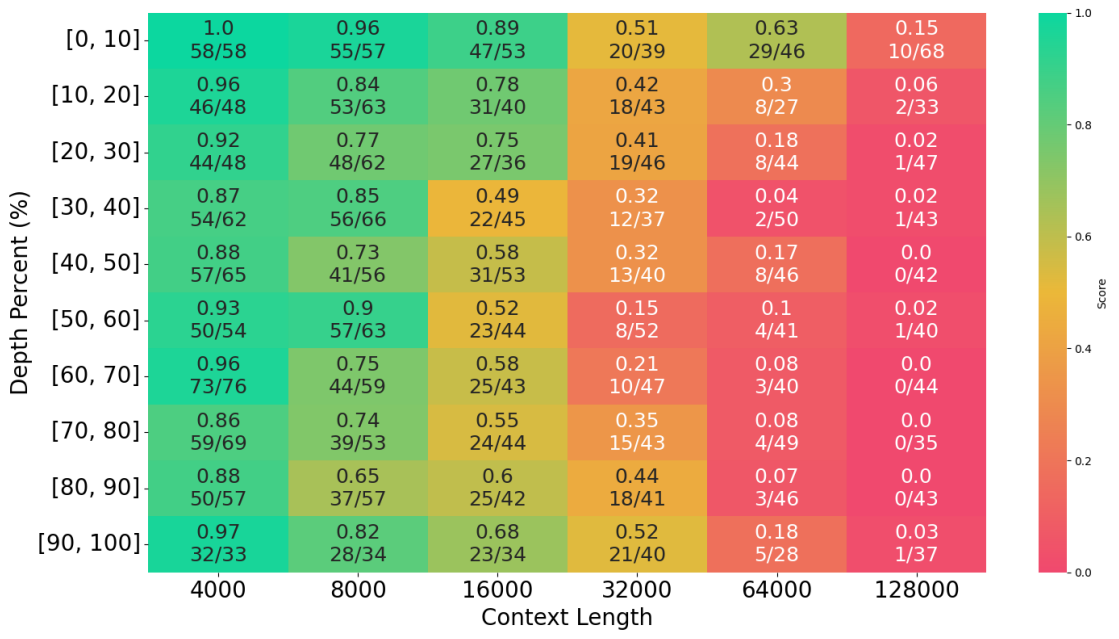

(a)

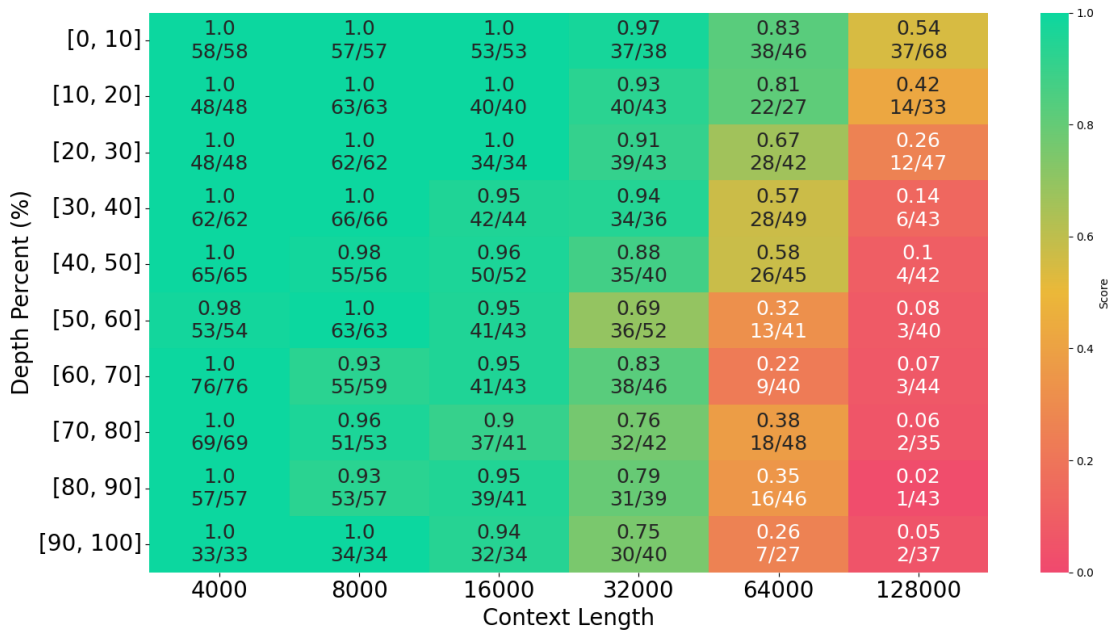

(b)

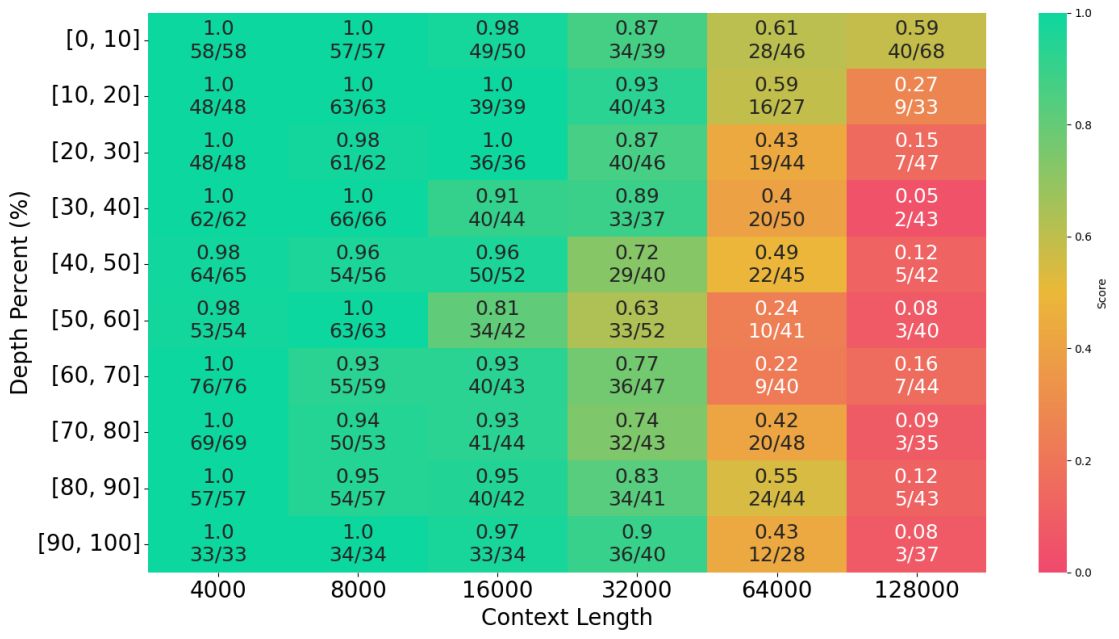

(c)

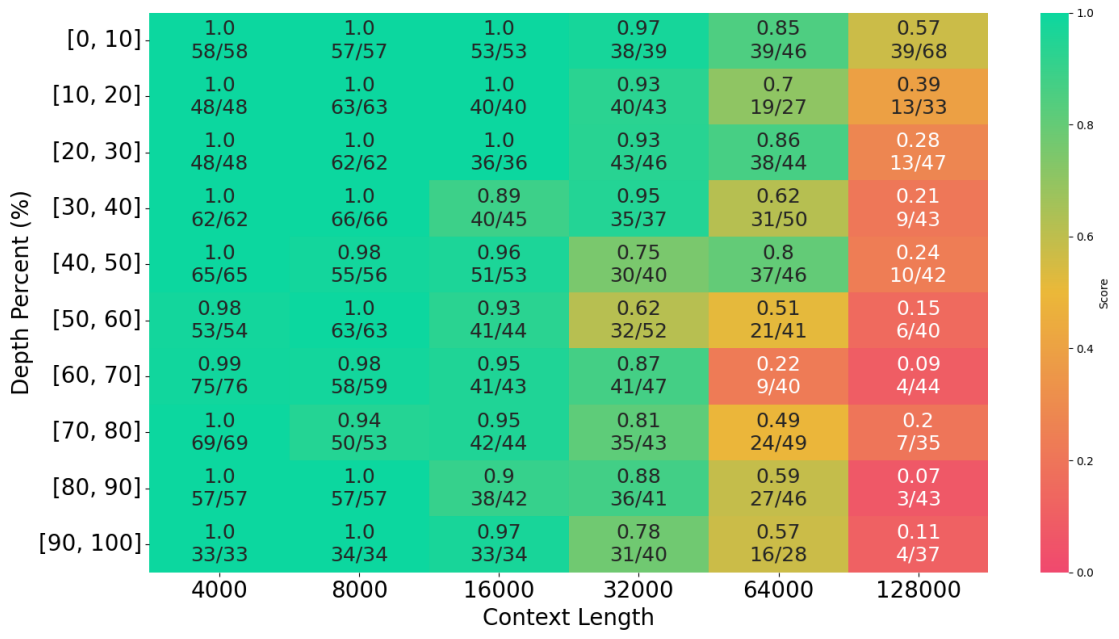

(d)

Figure 12: Prompt placement ablations using GPT-4-Turbo on "Company (C)" task with four configurations (Sec. 2.3): (a) Append (b) Prepend (c) Prepend+Append (d) OpenAI Best practices

| Model Name | Version | Decoding Strategy |
|---|---|---|
| GPT-4o | gpt-4o-2024-05-13 | Greedy |
| GPT-4-Turbo | gpt-4-turbo-2024-04-09 | Greedy |

Table 5: Details about model parameters

## A.9 Zero Needle

In Fig 13, we show the ability of GPT-4-Turbo to reject or refuse answering when ground-truth (or evidence) is absent from the context.

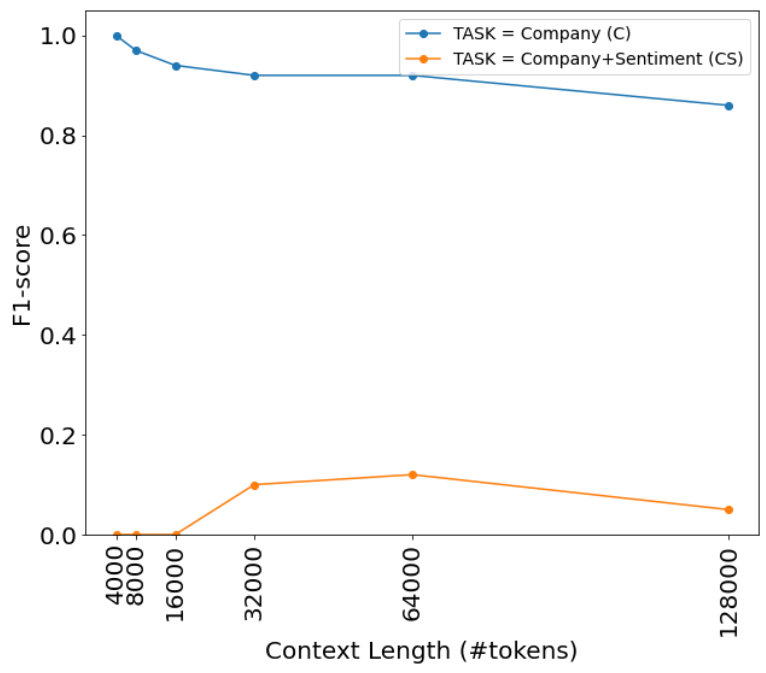

Figure 13: Performance of GPT-4-Turbo on zero-needle test records, ie. needle was absent from the haystack. Model is able to reject or refuse answering for easier tasks better than difficult tasks.